\documentclass{article} 
\usepackage{o2pd,times}
\usepackage{amsmath,amssymb,amsthm}
\usepackage{booktabs}
\usepackage{graphicx}
\usepackage{hyperref}
\usepackage{url}
\usepackage{xcolor}
\usepackage{colortbl}
\usepackage{enumitem}
\usepackage{caption}
\usepackage{titletoc}
\usepackage{wrapfig}
\usepackage{algorithm}
\usepackage{algpseudocode}
\usepackage{booktabs, multirow}
\definecolor{methodrow}{RGB}{237,244,252}

\title{O$^2$PD: Overlap-Gated On-Policy Distillation on Reliable Prefix}

\author{%
\begin{tabular}{c}
\bfseries
Shuaishuai Zu$^{1}$,
Jihao Zhao$^{1}$,
Pengfei Li$^{2}$,
Guanghuan Fang$^{2}$,
Ben Feng$^{2}$,
Yuhan Liu$^{2}$, \\
\bfseries
Biao Qin$^{*1}$,
Baochuan Yang$^{*2}$ \\[0.5em]
$^{1}$School of Information, Renmin University of China \\
$^{2}$Huawei Technologies Ltd.
\end{tabular}
}

\newcommand{\RKL}{\mathrm{RKL}}
\newcommand{\FKL}{\mathrm{FKL}}
\newcommand{\KL}{\mathrm{KL}}
\newcommand{\TopK}{\mathrm{TopK}}
\newsavebox{\takeawayboxcontent}

\newtheorem{lemma}{Lemma}
\newtheorem{proposition}{Proposition}

\iclrfinalcopy 
\begin{document}

\maketitle

\begin{abstract}
On-policy distillation (OPD) supplies dense token-level supervision by aligning a student to the teacher's next-token distribution on its own generated rollouts. Existing works mostly implement top-$k$ OPD, applying reverse Kullback-Leibler ($\RKL$) divergence between the teacher's and the student's renormalized distributions over teacher top-$k$ support.
However, the teacher scores student-generated rollouts that are inherently off-policy for itself, leaving the reliability of its next-token guidance unclear. Restricting the divergence to the teacher top-$k$ support also discards the remaining mass outside that support.
These gaps leave two questions implicit: 1) which tokens to align, and 2) how to align those tokens.
To answer these questions, we propose Overlap-gated On-Policy Distillation (O$^{2}$PD).
As for 1), we quantitatively analyze the reliability of teacher supervision and find that alignment on the student's trajectory prefix is more reliable than alignment on the full context. Therefore, our O$^{2}$PD is applied only to the 20\% prefix tokens of each student rollout.
As for 2), we theoretically characterize two problems of top-$k$ OPD: the \emph{Teacher-support Uncovering} and the \emph{Dominant-Mode Mismatch}. Based on these analyses, we propose two support strategies: the \emph{Reset Support} strategy attaches a reset atom to every token, pooling mass outside the teacher top-$k$, and the \emph{Support Convergence} strategy converges the support to the already aligned intersection. Our O$^{2}$PD employs the designed overlap-gated $\alpha$-divergence to interpolate between mode-covering and mode-seeking for per-token alignment, where the overlap is used as a gate to switch the support and adaptively modulate $\alpha$.
Extensive experiments on seven common benchmarks demonstrate the superiority of O$^{2}$PD, improving average performance by up to 5.29 points compared to OPD.
\end{abstract}

\section{Introduction}
\label{sec:intro}

Reinforcement learning with verifiable rewards (RLVR) improves reasoning by assigning advantages from outcome-level correctness~\cite{guo2025deepseekr1,shao2024deepseekmath,hu2025orz,zeng2025simplerl}. That signal is sparse along long trajectories: many tokens share one reward, and groups of equally correct or incorrect responses can yield vanishing advantages~\cite{yu2025dapo,liu2025drgrpo,cui2025entropy}. Knowledge distillation supplies denser targets by matching a student to a teacher's per-token distribution~\cite{hinton2015distilling}, but the usual recipe trains on teacher-generated sequences~\cite{kim2016sequence}. The student is then updated off-policy relative to its own sampler~\cite{gu2024minillm,agarwal2024gkd}. This introduces a distribution mismatch between training and those generated by the student itself.
On-policy distillation (OPD) reverses that sampling: the student generates the rollout trajectory, and the teacher scores the same rollout, so each token receives a dense, on-policy learning signal~\cite{agarwal2024gkd,gu2024minillm,lu2025onpolicy,yang2026gopd}.
The alignment process is conducted by minimizing the divergence between the teacher's and student's per-token distributions. For example, reverse Kullback-Leibler ($\RKL$) is a mode-seeking objective that can effectively capture the teacher’s dominant modes. Existing OPD methods mainly adopt top-$k$ OPD, minimizing the $\RKL$ on the teacher's top-$k$ support atoms for tractable cost. 

\begin{figure}[!t]
    \centering
    \includegraphics[width=0.95\linewidth]{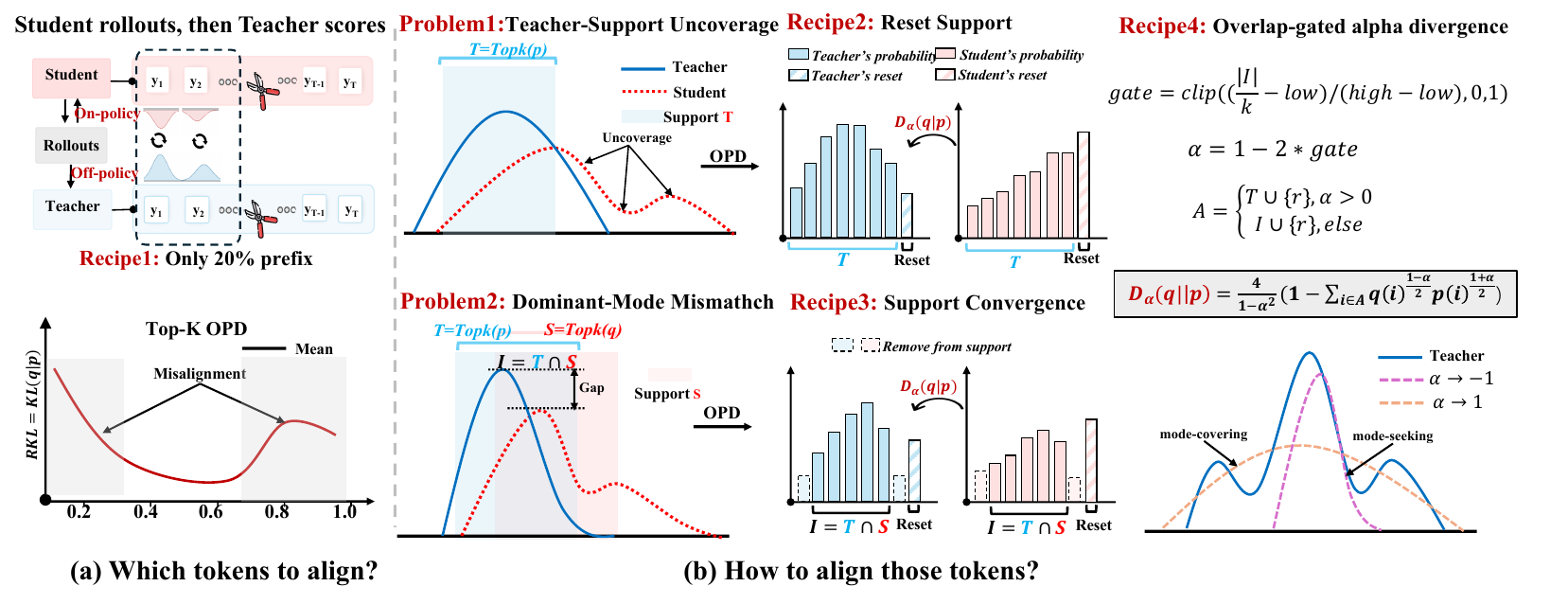}
    \caption{Framework of our O$^2$PD: (a)~Restricting alignment to the prefix, where teacher supervision is reliable. (b)~Overlap-gated $\alpha$-divergence with Reset Support and Support Convergence strategies.}
    \label{fig:model}
\end{figure}
However, the student-generated rollouts are inherently off-policy for the teacher, so the reliability of teacher supervision remains unclear~\cite{fu2026revisiting,zhou2026esr}.
Besides, the teacher's top-$k$ support atoms are just a subset of the full vocabulary~\cite{fu2026revisiting,anshumann2025sparse}. Restricting the alignment to this subset discards the remaining mass outside that support. Empirically, as the teacher-student overlap decreases, the student more frequently samples tokens outside the teacher’s top-$k$ support, where the teacher’s supervision is unreachable.
These gaps leave two questions implicit: 1) \emph{which tokens to align?}, and 2) \emph{how to align those tokens?}
Along these lines, we first identify a \emph{Cross-prefix Asymmetry phenomenon}: as student prefixes accumulate, the teacher's accuracy falls, while as teacher prefixes accumulate, the student's accuracy rises (\S\ref{sec:where}).
Furthermore, we theoretically characterize two problems of top-$k$ OPD (\S\ref{sec:how}): 1) \emph{Teacher-support Uncovering} happens because $\RKL$ on the teacher's top-$k$ support fails to pull the student onto teacher-only tokens; thus high-confidence teacher modes stay uncovered when overlap is low;
2) \emph{Dominant-Mode Mismatch} is that even the two top-$k$ supports already overlap well, the same objective only matches the renormalized distribution and fails to pin the teacher's mass on the shared atoms.

To address these issues, we propose an overlap-gated $\alpha$-divergence on refined support and reliable prefix for OPD, which we term O$^{2}$PD (shown in Figure~\ref{fig:model} ).
For \emph{which}, inspired by the Cross-prefix Asymmetry phenomenon, we only optimize the 20\% prefix tokens of student-generated rollouts to avoid the off-policy accumulation for the teacher.
For \emph{how}, we introduce two support strategies: the \emph{Reset Support} strategy attaches a reset atom to every token, pooling mass outside the teacher top-$k$, and the \emph{Support Convergence} strategy converges the support to the already aligned intersection.
Theoretical analysis shows that Forward KL ($\FKL$) with Reset Support supplies a first-order covering force on teacher-only atoms, and $\RKL$ with Support Convergence pins the teacher's mass on the shared atoms rather than leftover-renormalizing them. This motivates our design of overlap-gated $\alpha$-divergence, where overlap is used as a gate to switch the support strategies and adaptively modulate $\alpha$. For each token, $\alpha$ is mapped from the teacher-student overlap to control the interpolation between the mode-covering and the mode-seeking.
Our contributions are:
\begin{itemize}
\item We identify a \emph{Cross-prefix Asymmetry}: teacher supervision is more reliable on the prefix than on the suffix. We theoretically characterize two shortcomings of top-$k$ OPD, \emph{Teacher-support Uncovering} and \emph{Dominant-Mode Mismatch} (\S\ref{sec:where}, \S\ref{sec:how}).
\item We propose O$^{2}$PD, which applies overlap-gated $\alpha$-divergence on the reliable prefix tokens. The per-token overlap between teacher and student adaptively modulates $\alpha$ and switches \emph{Reset Support} and \emph{Support Convergence} strategies (\S\ref{sec:where}, \S\ref{sec:how}).
\item Extensive experiments on seven benchmarks show that O$^{2}$PD achieves state-of-the-art average performance among compared OPD methods, improving over OPD by up to $+5.29$ points across three teacher--student pairs.
\end{itemize}

\section{Related Work}

GKD, MiniLLM, and later OPD recipes minimize a token-level divergence loss on student-generated trajectories~\cite{gu2024minillm,yang2026gopd,ko2025distillm2}. The standard implementation is top-$k$ OPD~\citep{agarwal2024gkd,lu2025onpolicy,anshumann2025sparse}, which minimizes $\RKL$ over the teacher's top-$k$ atoms at every position.
At each token, the teacher and student induce next-token distributions $p$ and $q$ on vocabulary $V$. Write $T=\TopK(p)$, $S=\TopK(q)$, $I=T\cap S$, and overlap $\rho=\lvert I\rvert/k$ ( illustrated in Table~\ref{tab:support}). Top-$k$ OPD renormalizes both distributions on $T$ and minimizes
\begin{equation}
\label{eq:LT}
L_{T}
=\KL(q^{T}\Vert p^{T})
=\sum_{a\in T}q^{T}_a\log\frac{q^{T}_a}{p^{T}_a},
\end{equation}
where $q^{T}_a=q(a)/q(T)$ and $p^{T}_a=p(a)/p(T)$. Mass outside $T$ does not enter the loss.
A concurrent line of work studies when this recipe succeeds or fails. Because the teacher scores student-sampled rollouts, its next-token guidance can be noisy, unteachable, or mismatched with the student~\cite{fu2026revisiting,zhou2026esr,wang2026teachability,wang2026demystifying,yu2026tide,shao2026sure}. Li et al.\ show that teacher--student top-$k$ overlap tracks alignment-level and thinking-pattern consistency, and that high-overlap successful runs can be explained largely by overlap increasing~\cite{li2026rethinking}. 
\begin{wraptable}{r}{0.5\linewidth}
\caption{Supports at each token.}
\label{tab:support}
\centering
\small
\begin{tabular}{ll}
\toprule
Set & Role \\
\midrule
$T=\TopK(p)$ & teacher support \\
$S=\TopK(q)$ & student support \\
$I=T\cap S$ & already aligned atoms \\
$T\setminus S$ & teacher modes still to cover \\
$S\setminus T$ & student-only modes to suppress \\
\bottomrule
\end{tabular}
\end{wraptable}
Other methods intervene on which tokens to optimize and how to lessen the overlap mismatch. For example, TIP selects tokens with high student entropy, and tokens with low student entropy plus high teacher--student divergence, where the student is overconfident and wrong~\cite{xu2026tip}. ESR argues that early tokens do problem framing and strategic planning, and truncates the rollout to a fixed first $N$ tokens~\cite{zhou2026esr}. EOPD keeps the on-policy trajectory and mixes $\FKL$ into $\RKL$ on high teacher-entropy tokens to avoid diversity collapse~\cite{jin2026eopd}. 
However, these token filters are heuristic, ESR's cutoff is a fixed short horizon that does not scale with response length, and EOPD changes the alignment objective without using overlap to decide covering versus pinning. O$^{2}$PD takes both decisions together: a length-scaled prefix, and overlap-gated $\alpha$-divergence with Reset Support and Support Convergence for better alignment.

\section{Where Should Distillation Concentrate?}
\label{sec:where}

\subsection{Cross-prefix Asymmetry phenomenon}
\begin{figure*}[!t]
    \centering
    \includegraphics[width=\linewidth]{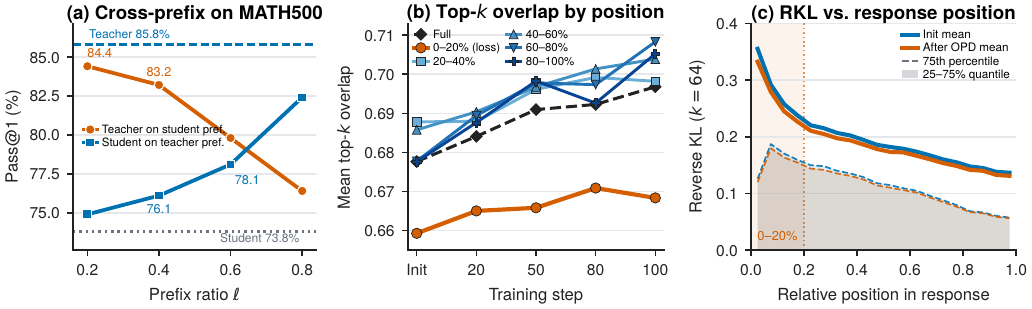}
    \caption{(a)~Cross-prefix pass@1 versus prefix ratio $\ell$ on MATH500. (b)~Mean top-$k$ overlap by relative-position bin when $L_T$ is applied only to the first $20\%$ of each student trajectory. (c)~Top-$k$ OPD loss $L_T$ versus relative position at initialization and after full-trajectory training.}
    \label{fig:where}
\end{figure*}

\textbf{Student prefixes are off-policy for the teacher.}
To test whether teacher supervision remains reliable on student-generated prefixes, we run a cross-prefix completion protocol on MATH500 with a Qwen3-1.7B student and a Qwen3-4B teacher. When each model generates the entire response, the student and teacher obtain $73.8\%$ and $85.8\%$ pass@1. We then vary a prefix ratio $\ell$: the student continues from a teacher-generated prefix of relative length $\ell$, and the teacher continues from a student-generated prefix of the same length. Fig.~\ref{fig:where} (a) shows an opposite dependence on $\ell$. Student pass@1 increases as the teacher prefix lengthens; teacher pass@1 decreases as the student prefix lengthens, and the decline grows with $\ell$. 

\textbf{Aligning prefix raises suffix overlap.}
To test whether aligning only those early tokens can raise teacher--student overlap on the rest of the trajectory, we use the same Qwen3-1.7B student and Qwen3-4B teacher and train top-$k$ OPD on DAPO-MATH-17K with the loss applied only to the first $20\%$ of each rollout. As shown in Fig.~\ref{fig:where} (b), we compute mean top-$k$ overlap between the student and the teacher, grouped by relative-position bin.
Overlap on the untrained suffix bins rises together with the trained prefix, so the prefix-only objective improves alignment for suffix tokens. The $0$--$20\%$ bin remains the lowest throughout, so the prefix is still the hardest region to match. \S\ref{sec:ablate} further verifies that aligning on a prefix fraction exceeds full-sequence alignment, and concentrating on the prefix outperforms aligning the suffix.

\textbf{Misalignment concentrates on the prefix.}
As shown in Fig.~\ref{fig:where} (c), we plot mean $L_T$ against relative position at initialization and after full-trajectory top-$k$ OPD.
Mean $\RKL$ is highest near the start of the response and declines with relative position at both checkpoints. Training lowers the curve along the trajectory, so teacher-student mismatch is alleviated, yet the prefix remains the region of largest mismatch. 



\subsection{Recipe1: Prefix-only Alignment}
Based on this analysis, the prefix is both the most reliable source of teacher supervision and the region of largest remaining mismatch, so these are the tokens most worth aligning. Prefix-only training still raises overlap on the untrained suffix, so withholding suffix tokens from the loss does not leave the suffix unaligned. We therefore replace full-trajectory $L_T$ by the Where objective $\mathcal{L}_{\mathrm{where}}^{(\ell)}$:
\begin{equation}
\label{eq:prefix}
\mathcal{L}_{\mathrm{where}}^{(\ell)}
=\mathbb{E}_{y\sim\pi_s(\cdot\mid x)}
\sum_{i=1}^{|y|}
\mathbf{1}\bigl[i\le \ell\,|y|\bigr]
L_T^{(i)}.
\end{equation}
Suffix tokens are sampled with the student but do not enter the optimization. Because the window is a fraction of $|y|$, the number of aligned tokens scales with response length.

\section{How to Align Prefix Tokens}
\label{sec:how}


\subsection{Teacher-support Uncovering and Dominant-Mode Mismatch}
\label{sec:failures}

We theoretically characterize two shortcomings of top-$k$ OPD.
The first failure is Teacher-support Uncovering: minimizing $\RKL$ on $T$ involves gradient coordinates only on $T$, so it therefore assigns no gradient to student-only atoms $S\setminus T$, and its first-order force on teacher-only atoms in $T\setminus S$ is multiplied by student mass.

\begin{lemma}[Support restriction]
\label{lem:support-restriction}
$L_T=\KL(q^{T}\Vert p^{T})$ depends on the student logits only through $\{z_a:a\in T\}$. Consequently $\partial L_T/\partial z_a=0$ for every $a\notin T$. In particular, $\RKL$ on $T$ has no coordinate on $S\setminus T$.
\end{lemma}

On the simplex over $T$, the coordinate gradient of $L_T$ is proportional to student mass $q^{T}_a$, so the first-order force on $T\setminus S$ is weak whenever that mass is small (Appendix~\ref{app:rkl-grad}).

\begin{proposition}[Covering channel]
\label{prop:covering}
Assume $I\neq\emptyset$, $q(I)>0$, and $q(T\setminus S)=0$. Then $q(T)=q(I)$, $q^{T}$ vanishes on $T\setminus S$, and
\[
\frac{\partial L_T}{\partial\zeta_a}=0\qquad\text{for all }a\in T\setminus S.
\]
In this case $L_T$ is determined by $q^{I}$ only, so the first-order update lives only on already aligned atoms. The covering channel on $T\setminus S$ is off, and Lemma~\ref{lem:support-restriction} additionally gives $\partial L_T/\partial z_a=0$ for every $a\in S\setminus T$. When $q(T\setminus S)$ is small but nonzero, the covering gradient remains proportional to $q^{T}_a$, so the first-order force on $T\setminus S$ is weak rather than identically zero.
\end{proposition}

\begin{figure}[!htbp]
    \centering
    \includegraphics[width=\linewidth]{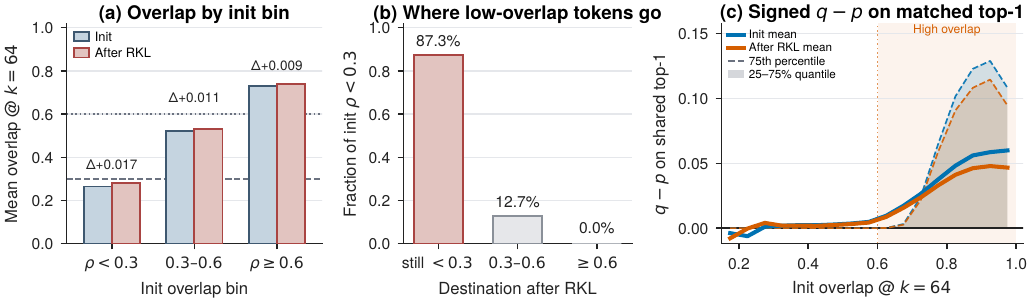}
    \caption{ (a)~Mean overlap by initial-overlap bin before and after training. (b)~Destination of tokens with initial $\rho<0.3$. (c)~Signed gap $q-p$ on the shared top-$1$.}
    \label{fig:lowov}
\end{figure}

Proposition~\ref{prop:covering} is why $L_T$ does not pull $T\setminus S$ into $S$. Fig.~\ref{fig:lowov} checks this on prefix tokens with initial overlap $\rho<0.3$: mean overlap moves only from $0.265$ to $0.282$ after top-$k$ OPD training, $87.3\%$ of the cohort remain below $0.3$, and none reach $\ge 0.6$.

The second failure is Dominant-Mode Mismatch: minimizing $\RKL$ on the full-vocabulary is associated with $q=p$ as a mode-seeking target. Top-$k$ OPD is not full-vocabulary OPD: its loss $L_T$ lives on the renormalized simplex of $T$.

\begin{proposition}[Renormalized match on $T$]
\label{prop:sampled-min}
On the relative interior $\{q^{T}:q(T)>0\}$, every minimizer of $L_T$ satisfies $q^{T}=p^{T}$. This matches the teacher only as a distribution on $T$. It does not pin $q(T)=p(T)$ and does not constrain $V\setminus T$, so it is not $q=p$ on $V$.
\end{proposition}

Remaining mode-mass gaps are therefore a shape mismatch on the support $T$, not evidence that the student is insufficiently mode-seeking. Fig.~\ref{fig:lowov} (c) checks tokens that already share the teacher's top-$1$: the signed gap $q-p$ on that mode stays positive after top-$k$ OPD and grows with initial overlap. High overlap does not mean the student has matched the teacher's dominant-mode mass. Proofs of Lemma~\ref{lem:support-restriction} and Propositions~\ref{prop:covering}--\ref{prop:sampled-min} are in Appendix~\ref{app:proofs}.

\subsection{Recipe2: a reset atom on every token (Reset Support)}
\label{sec:fkl-rest}
Teacher-support Uncovering can be handled by our Reset Support strategy, which attaches a reset atom to every token. The reset atom pools the mass outside $T$: teacher reset mass is $1-p(T)$ and student reset mass is $1-q(T)$. On the simplex over $T\cup\{r\}$,
\begin{equation}
\pi_p(a)=p(a)\ (a\in T), \pi_p(r)=1-p(T),
\pi_q(a)=q(a)\ (a\in T), \pi_q(r)=1-q(T).
\end{equation}
$\FKL$ on this simplex is $L_F=\KL(\pi_p\Vert\pi_q)$. Without a reset atom, $\FKL$ opens only a covering channel on $T\setminus S$. The reset coordinate additionally supplies a pooled suppression channel for $S\setminus T\subset V\setminus T$.

\begin{proposition}[FKL covering]
\label{prop:fkl-cover}
On simplex coordinates $\pi_q=\mathrm{softmax}(\zeta)$,
\[
\frac{\partial L_F}{\partial\zeta_a}=\pi_q(a)-\pi_p(a).
\]
If $a\in T\setminus S$, $\pi_p(a)>0$ and $\pi_q(a)=0$, then $\partial L_F/\partial\zeta_a=-\pi_p(a)<0$, so a descent step raises $\zeta_a$. Teacher mass on $T\setminus S$ supplies covering force without requiring the student to already place mass there. The reset coordinate yields $\partial L_F/\partial\zeta_r=q_r-p_r$ with $q_r=1-q(T)$ and $p_r=1-p(T)$. When $T\setminus S$ is large the student typically places substantial mass outside $T$, so $q_r>p_r$ and a descent step lowers the reset coordinate, a pooled suppression of $S\setminus T$. If $p(T\setminus S)>0$, then $L_F\to+\infty$ as $q^{T}|_{T\setminus S}\to 0$, whereas $L_T$ remains finite.
\end{proposition}

The reason to use $\FKL$ when $T\setminus S$ is large: the covering channel stays open when student mass on $T\setminus S$ is near zero. $\RKL$ on the same support still multiplies those coordinates by $q(a)$.

\subsection{Recipe3: shrink the support, then $\RKL$ (Support Convergence)}
\label{sec:i-rest}

Dominant-Mode Mismatch means that minimizing $\RKL$ on $T$ matches only $q^{T}=p^{T}$. Attaching a reset atom to $T$ makes the unconstrained $\RKL$ minimizer $\pi_q=\pi_p$ on $T\cup\{r\}$. On $T\cup\{r\}$ this is $q|_T=p|_T$, which pins $c=q(T)/p(T)=1$ and closes the scale gap in Proposition~\ref{prop:sampled-min}. Reset Support is therefore already enough for the unconstrained problem.
As overlap rises, $T\setminus S$ shrinks and the job becomes mode-seeking on the already shared atoms $I$. The interiors of $T\cup\{r\}$ and $I\cup\{r\}$ coincide only when $T=I$. Leftover atoms $T\setminus I$ still carry teacher mass but little student mass. Proposition~\ref{prop:covering} keeps the $\RKL$ factor on those atoms proportional to $q(a)$, so leftover student mass $\varepsilon=q(T\setminus I)$ typically remains below $p(T\setminus I)$ rather than rising to the unconstrained leftover target. On this slice $\RKL$ on $T\cup\{r\}$ still renormalizes $I$ by leftover teacher mass.
Support Convergence strategy is therefore proposed to restrict the support to $I$. Leftover teacher mass is folded into the reset atom,
$
\pi_q(a)=q(a)\ (a\in I), \pi_q(r)=1-q(I),
\pi_p(a)=p(a)\ (a\in I), \pi_p(r)=1-p(I),
$
and $L_{I,r}=\KL(\pi_q\Vert\pi_p)$.

\begin{proposition}[Leftover renormalization]
\label{prop:face-renorm}
Assume $I\neq\emptyset$, $T\setminus I\neq\emptyset$, $q(I)>0$, and $p(T)<1$. Write $\varepsilon=q(T\setminus I)\in[0,1)$ and freeze the leftover coordinates. Then $\RKL$ on $T\cup\{r\}$, as a function of the remaining mass $1-\varepsilon$ on $I\cup\{r\}$, is uniquely minimized at
\[
q(a)=\frac{(1-\varepsilon)\,p(a)}{1-p(T\setminus I)}\quad(a\in I),\qquad
1-q(I)-\varepsilon=\frac{(1-\varepsilon)\,(1-p(T))}{1-p(T\setminus I)}.
\]
Hence $q(a)>p(a)$ on $I$ if and only if $\varepsilon<p(T\setminus I)$. The uncovered face $\varepsilon=0$ is the special case $q(a)=p(a)/(1-p(T\setminus I))$. Minimizing $\RKL$ on $I\cup\{r\}$ does not include $T\setminus I$; its interior minimizer is $q|_I=p|_I$ and $1-q(I)=1-p(I)$. Leftover teacher mass therefore renormalizes the aligned atoms under $T\cup\{r\}$ whenever student leftover mass is smaller than teacher leftover mass, and is folded into the reset atom under $I\cup\{r\}$.
\end{proposition}
Proposition~\ref{prop:face-renorm} is why mode-seeking $\RKL$ uses support $I\cup\{r\}$: when leftover mass remains below leftover teacher mass, $T\cup\{r\}$ still overshoots $p$ on $I$, not only on the exact-zero face. Proofs of Propositions~\ref{prop:fkl-cover} and~\ref{prop:face-renorm} are in Appendix~\ref{app:proofs}.

\subsection{Recipe4: Overlap-gated $\alpha$-divergence}
\label{sec:algo}

Low overlap needs the covering force of $\FKL$ on $T\cup\{r\}$ (Proposition~\ref{prop:fkl-cover}); high overlap needs $\RKL$ on $I\cup\{r\}$, where $T\cup\{r\}$ would still leftover-renormalize $I$ (Proposition~\ref{prop:face-renorm}). Overlap $\rho$ selects which of those two jobs a prefix token presents.
We therefore propose overlap-gated $\alpha$-divergence: $\rho$ interpolates those two regimes and switches the support strategies.
Let $\pi_q$ and $\pi_p$ be the student and teacher distributions on the simplex over the current support $A$, either $T\cup\{r\}$ or $I\cup\{r\}$. For $\alpha\in(-1,1)$, the $\alpha$-divergence is
\begin{equation}
\label{eq:alpha-div}
D_{\alpha}(\pi_q\Vert\pi_p)
=\frac{4}{1-\alpha^{2}}
\Biggl(1-\sum_{i\in A}
\pi_q(i)^{\frac{1-\alpha}{2}}
\pi_p(i)^{\frac{1+\alpha}{2}}\Biggr).
\end{equation}
The endpoints recover the two KL directions used above: $D_{\alpha}\to\KL(\pi_p\Vert\pi_q)$ as $\alpha\to 1$, and $D_{\alpha}\to\KL(\pi_q\Vert\pi_p)$ as $\alpha\to -1$. Appendix~\ref{app:alpha-lim} verifies the two limits for \eqref{eq:alpha-div}. Intermediate $\alpha$ interpolates those two vector fields rather than adding a second loss.
On each update, let $\mathcal{P}$ be the prefix tokens that enter the loss. Set
$
\mathrm{low}=\min_{i\in\mathcal{P}}\rho_i,
\mathrm{high}=\max_{i\in\mathcal{P}}\rho_i,
$
and let $\rho_{0.5}$ be the median percentile of $\{\rho_i:i\in\mathcal{P}\}$. Define
\begin{equation}
\mathrm{gate}=\mathrm{clip}\bigl((\rho-\mathrm{low})/(\mathrm{high}-\mathrm{low}),0,1\bigr),
\qquad
\alpha=1-2\cdot\mathrm{gate}.
\end{equation}
The support switches at the median: $T\cup\{r\}$ when $\rho<\rho_{0.5}$, and $I\cup\{r\}$ when $\rho\ge\rho_{0.5}$.
We therefore propose the overlap-gated $\alpha$-divergence for adaptively aligning each token via $\rho$:
\begin{equation}
\label{eq:how}
\mathcal{L}_{\mathrm{how}}
=\mathbb{E}_{y\sim\pi_s(\cdot\mid x)}
\sum_{i=1}^{|y|}
D_{\alpha_i}\bigl(\pi_q^{(i)}\Vert\pi_p^{(i)}\bigr),
\end{equation}
where $\rho_i$ determines both $\alpha_i$ and the support of $D_{\alpha_i}$ as above.
The full objective puts the Where prefix window on this How divergence. With prefix fraction $\ell=0.2$ from \S\ref{sec:where},
\begin{equation}
\label{eq:o2pd}
\mathcal{L}
=\mathbb{E}_{y\sim\pi_s(\cdot\mid x)}
\sum_{i=1}^{|y|}
\mathbf{1}\bigl[i\le \ell\,|y|\bigr]
\,D_{\alpha_i}\bigl(\pi_q^{(i)}\Vert\pi_p^{(i)}\bigr).
\end{equation}
Suffix tokens are sampled with the student but do not enter the loss.

\section{Experiments}
\label{sec:exp}

\subsection{Setup}
\label{sec:setup}

\textbf{Models and training dataset.}
We use three student--teacher pairs (Table~\ref{tab:distill-avg}). Qwen3-1.7B distilled from Qwen3-4B~\cite{yang2025qwen3} is a same-family distillation with a larger teacher. DSR1-1.5B distilled from JustRL-1.5B~\cite{guo2025deepseekr1,he2025justrl} is a same-size distillation on long trajectories. Qwen3-1.7B distilled from Qwen3-4B-Non-Thinking-RL (NTRL)~\cite{yang2025qwen3} keeps the first student and changes only the teacher, isolating teacher strength.
The training set is DAPO-MATH-17K with no additional filtering~\cite{yu2025dapo}. The compared baselines are: OPD, ESR~\cite{zhou2026esr}, EOPD~\cite{jin2026eopd}, and TIP~\cite{xu2026tip}. OPD is the default top-$k$ recipe. ESR, EOPD, and TIP are trained with the configurations from their papers. 

\textbf{Evaluation.}
For fair comparison, we follow previous evaluation pipeline~\cite{fu2026revisiting,lu2025onpolicy} and evaluate on seven common benchmarks: AIME24, AIME25, AIME26, HMMT25, Minerva~\cite{lewkowycz2022minerva}, Olympiad~\cite{he2024olympiadbench}, and MATH500~\cite{hendrycks2021math,lightman2024verify}. AIME24/25/26 and HMMT25 have few problems and report Avg@8; Minerva, Olympiad, and MATH500 have many more and report pass@1 from a single sample. Reported scores use the best checkpoint of each method. 

\textbf{Implementation.}
All methods are trained in VERL~\cite{sheng2025hybridflow}. OPD and O$^{2}$PD use distillation top-$k$ size $k=16$. O$^{2}$PD applies the loss to a prefix fraction $\ell=0.2$, attaches a reset atom, maps $\alpha$ from the prefix overlap min and max, and switches named support at the median of prefix overlap: $T\cup\{r\}$ below the median and $I\cup\{r\}$ at or above it.
Further training hyperparameters are in Appendix~\ref{app:hyper}, and evaluation details in Appendix~\ref{app:eval}.

\setlength{\textfloatsep}{8pt plus 2pt minus 2pt}
\begin{table*}[!b]
\centering
\caption{Performance (\%) for three OPD student--teacher pairs. Len is the mean generated length in tokens.
Bold marks the best distilled method in each column within a pair; underline marks the second best.
$\Delta$ vs.\ OPD is O$^{2}$PD minus OPD.}
\label{tab:distill-avg}
\small
\setlength{\tabcolsep}{2pt}
\renewcommand{\arraystretch}{0.9}
\setlength{\aboverulesep}{0.6ex}
\setlength{\belowrulesep}{0.4ex}
\begin{tabular}{@{}l ccccccc cc@{}}
\toprule
Method & AIME24 & AIME25 & AIME26 & HMMT25 & Minerva & Olympiad & MATH500 & Len & Avg \\
\midrule
\multicolumn{10}{c}{Qwen3-1.7B (student) $\leftarrow$ Qwen3-4B (teacher)} \\
\midrule
Qwen3-1.7B & 8.33 & 12.08 & 5.83 & 6.25 & 20.59 & 36.74 & 73.8 & 3049 & 23.37 \\
Qwen3-4B & 23.33 & 31.67 & 21.67 & 15.00 & 31.99 & 45.48 & 85.8 & 4127 & 36.42 \\
\midrule
OPD & 13.75 & 14.17 & \underline{7.08} & 5.83 & \underline{23.53} & 36.00 & 74.0 & 5033 & 24.91 \\
ESR & 13.75 & 14.17 & \underline{7.08} & \textbf{8.33} & 21.32 & 38.37 & 74.2 & 4480 & 25.32 \\
EOPD & \underline{14.17} & 12.50 & \textbf{11.67} & 5.00 & \textbf{23.90} & 38.37 & 74.6 & 5110 & \underline{25.74} \\
TIP & 12.92 & \underline{14.58} & 6.67 & \underline{6.67} & 21.32 & \textbf{39.70} & \textbf{76.8} & 5377 & 25.52 \\
\midrule
\rowcolor{methodrow}
O$^{2}$PD & \textbf{15.42} & \textbf{15.00} & \textbf{11.67} & \underline{6.67} & \textbf{23.90} & \underline{38.96} & \underline{76.0} & 4684 & \textbf{26.80} \\
\rowcolor{methodrow}
$\Delta$ vs.\ OPD & $+1.67$ & $+0.83$ & $+4.59$ & $+0.84$ & $+0.37$ & $+2.96$ & $+2.0$ & $--$ & $+1.89$ \\
\midrule
\multicolumn{10}{c}{DSR1-1.5B (student) $\leftarrow$ JustRL-1.5B (teacher)} \\
\midrule
DSR1-1.5B & 28.75 & 23.75 & 21.25 & 12.08 & 18.38 & 28.00 & 84.4 & 15627 & 30.94 \\
JustRL-1.5B & 50.83 & 40.00 & 35.83 & 22.08 & 27.94 & 42.96 & 91.4 & 9308 & 44.43 \\
\midrule
OPD & 42.92 & 32.08 & 31.67 & 16.67 & 16.91 & 18.67 & 83.8 & 12241 & 34.67 \\
ESR & 36.67 & 30.00 & 27.50 & 15.83 & \underline{20.22} & 25.04 & 83.4 & 12376 & 34.09 \\
EOPD & 42.50 & \textbf{34.17} & 23.75 & 17.50 & 18.38 & \underline{29.48} & 86.0 & 11677 & 35.97 \\
TIP & \underline{45.83} & 31.67 & \textbf{31.67} & \underline{18.75} & 17.65 & 25.63 & \underline{86.6} & 11552 & \underline{36.83} \\
\midrule
\rowcolor{methodrow}
O$^{2}$PD & \textbf{46.67} & \underline{33.33} & \underline{30.42} & \textbf{20.00} & \textbf{23.16} & \textbf{33.78} & \textbf{89.0} & 10875 & \textbf{39.48} \\
\rowcolor{methodrow}
$\Delta$ vs.\ OPD & $+3.75$ & $+1.25$ & $-1.25$ & $+3.33$ & $+6.25$ & $+15.11$ & $+5.2$ & $--$ & $+4.81$ \\
\midrule
\multicolumn{10}{c}{Qwen3-1.7B (student) $\leftarrow$ Qwen3-4B-NTRL (teacher)} \\
\midrule
Qwen3-1.7B & 8.33 & 12.08 & 5.83 & 6.25 & 20.59 & 36.74 & 73.8 & 3049 & 23.37 \\
Qwen3-4B-NTRL & 70.00 & 63.33 & 65.00 & 38.75 & 39.34 & 64.44 & 94.8 & 11751 & 62.24 \\
\midrule
OPD & 45.42 & 40.42 & 37.08 & 21.25 & 31.62 & 51.85 & 89.4 & 19129 & 45.29 \\
ESR & 46.67 & 37.92 & 42.50 & 22.92 & 33.09 & \underline{54.22} & 90.2 & 13893 & 46.79 \\
EOPD & \underline{52.08} & 39.17 & 40.83 & \underline{23.33} & \underline{34.56} & 53.19 & \underline{90.2} & 17696 & 47.62 \\
TIP & 50.42 & \underline{44.58} & \underline{43.33} & 22.92 & 32.72 & 52.74 & 89.8 & 19036 & \underline{48.07} \\
\midrule
\rowcolor{methodrow}
O$^{2}$PD & \textbf{54.17} & \textbf{45.83} & \textbf{44.58} & \textbf{25.83} & \textbf{36.76} & \textbf{55.70} & \textbf{91.2} & 19631 & \textbf{50.58} \\
\rowcolor{methodrow}
$\Delta$ vs.\ OPD & $+8.75$ & $+5.41$ & $+7.50$ & $+4.58$ & $+5.14$ & $+3.85$ & $+1.8$ & $--$ & $+5.29$ \\
\bottomrule
\multicolumn{10}{p{\dimexpr\textwidth-2\tabcolsep\relax}}{\footnotesize
}
\end{tabular}
\end{table*}

\subsection{Main distillation}
\label{sec:main}

Table~\ref{tab:distill-avg} reports seven-suite Avg for the three student--teacher pairs. O$^{2}$PD attains $26.80$ with the Qwen3-4B teacher, $39.48$ with JustRL-1.5B, and $50.58$ with NTRL. On every pair it exceeds ESR, EOPD, and TIP on Avg, and it improves over OPD by $+1.89$, $+4.81$, and $+5.29$.
TIP and ESR raise Avg performance over OPD on all three teacher-student pairs, by up to $+3.63$ and $+2.36$. This reveals that optimizing a subset of tokens is more effective than optimizing all tokens. However, ESR's Avg gains remain small, and on some benchmarks ESR falls behind OPD, which demonstrates that dynamically adjusting the number of selected tokens may be necessary on harder problems. EOPD also raises Avg over OPD, by up to $+3.18$, by mixing $\FKL$ into $\RKL$ on high-entropy teacher tokens, showing that $\RKL$ alone is insufficient and benefits from an $\FKL$ covering channel.
Instead of applying complex token-selective strategies, we optimize a fixed fraction of each trajectory's prefix, so the number of aligned tokens scales with the response length. We then use overlap-gated $\alpha$-divergence to combine $\FKL$ covering with $\RKL$ mode-seeking and switch the support with overlap. This yields a further Avg gain over ESR, EOPD, and TIP on all three teacher-student pairs.

\subsection{Discussion}
\label{sec:ablate}

\begin{figure}[!ht]
    \centering
    \includegraphics[width=1\linewidth]{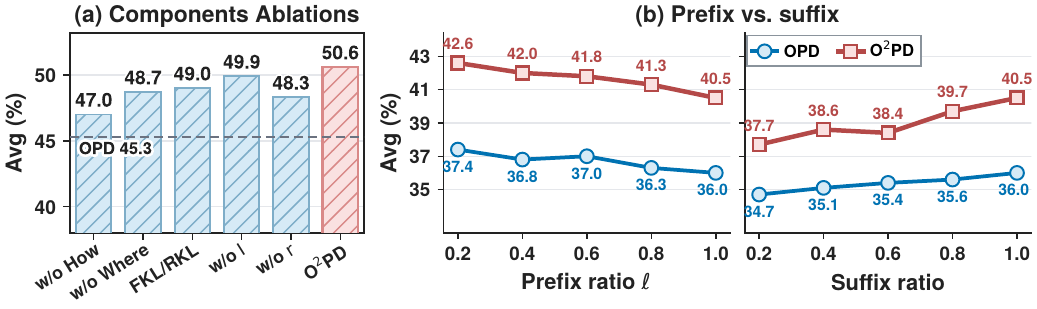}
    \caption{(a) Method-component ablation studies (\%). Each bar removes one component of O$^{2}$PD. (b) Prefix versus suffix performance. Avg is the mean of Avg@8 on AIME24, AIME25, AIME26, and HMMT25.}
    \label{fig:ablation}
\end{figure}

\textbf{Method components.} \label{sec:ablate-components}We ablate five variants (Fig.~\ref{fig:ablation}). \emph{w/o How} applies the prefix restriction to OPD. \emph{w/o Where} applies overlap-gated \(\alpha\)-divergence with Reset Support and Support Convergence to the full trajectory. \emph{$\FKL$/$\RKL$} uses \(\FKL\) on \(T\cup\{r\}\) below the prefix overlap median and \(\RKL\) on \(I\cup\{r\}\) at or above it. \emph{w/o \(I\)} keeps the prefix and runs overlap-gated \(\alpha\)-divergence on \(T\cup\{r\}\) only. \emph{w/o \(r\)} keeps the prefix and runs overlap-gated \(\alpha\)-divergence on the teacher top-\(k\) \(T\), without the reset atom.

On the Qwen3-1.7B student with the NTRL teacher, all five variants lie between OPD and full O$^{2}$PD on seven-suite Avg (Fig.~\ref{fig:ablation} (a)). \emph{w/o How} exceeds full-trajectory OPD, so the prefix restriction alone beats applying the loss to every token. \emph{w/o Where} is higher, so How on the full trajectory contributes more than the prefix restriction alone. Dropping either block leaves a gap to full O$^{2}$PD.
\begin{wrapfigure}[12]{r}{0.5\linewidth}
\centering
    \includegraphics[width=\linewidth]{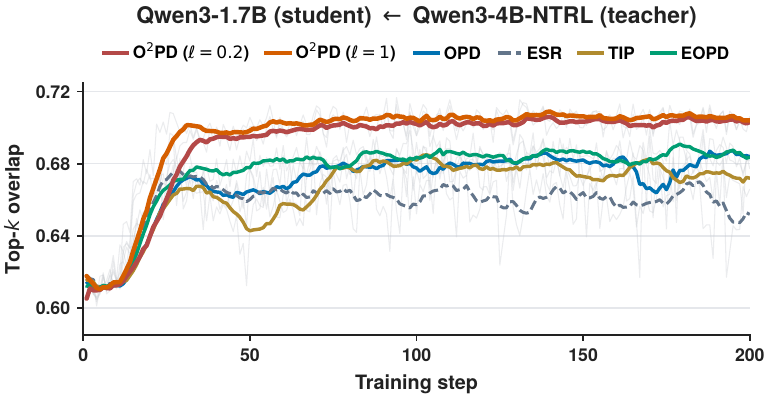}
    \caption{Mean top-$k$ overlap trend.}
    \label{fig:overlap}
\end{wrapfigure}
Within How, \emph{w/o \(r\)} is the largest single-component drop; the reset atom carries the covering channel. With the support already switching with overlap, \emph{$\FKL$/$\RKL$} stays below both \emph{w/o \(I\)} and full O$^{2}$PD: continuous \(\alpha\) interpolates the two vector fields rather than adding a second loss. Converging the support to \(I\cup\{r\}\) as overlap rises is the smallest increment.

\textbf{Prefix versus suffix.}\label{sec:ablate-prefix-suffix} To test whether the teacher provides effective supervision on prefix tokens relative to suffix tokens and to the full rollout, we vary where the loss sits on the full trajectory for both OPD and O$^{2}$PD.
The prefix sweep applies the loss to the first fraction \(\ell\) of each student response; the suffix sweep applies it to the last fraction. The two methods share the full-trajectory endpoint \(\ell=1\).
Both methods are higher at \(\ell=0.2\) than at full-trajectory training (Fig.~\ref{fig:ablation} (b)). Optimizing the suffix pulls down that distillation improvement.
At a matched budget of \(0.2\), the prefix outperforms the suffix for both methods, and suffix-only training falls below the full trajectory. The teacher therefore provides effective supervision on prefix tokens. O$^{2}$PD remains above OPD at every measured endpoint, and the best point is prefix \(\ell=0.2\) with the How recipes.

\textbf{Overlap dynamics.}\label{sec:ablate-overlap} To test whether O$^{2}$PD raises teacher--student alignment level, measured by mean top-$k$ overlap, and whether the prefix fraction \(\ell=0.2\) still attains the alignment of full-trajectory distillation, we track that overlap over training on the Qwen3-1.7B student with the NTRL teacher for O$^{2}$PD at \(\ell=0.2\) and \(\ell=1\), together with OPD, ESR, TIP, and EOPD (Fig.~\ref{fig:overlap}).
All methods start near \(0.61\), rise in the first few steps, and then plateau through step \(200\). Distillation under O$^{2}$PD at \(\ell=1\) rises faster than at \(\ell=0.2\); the prefix run then catches up, and both finish near \(0.70\). That level is above OPD, EOPD, and TIP near \(0.68\), and above ESR. Prefix-only distillation therefore delays alignment and does not lower the level O$^{2}$PD attains. 

\begin{wrapfigure}[17]{r}{0.5\linewidth}
\centering
    \includegraphics[width=\linewidth]{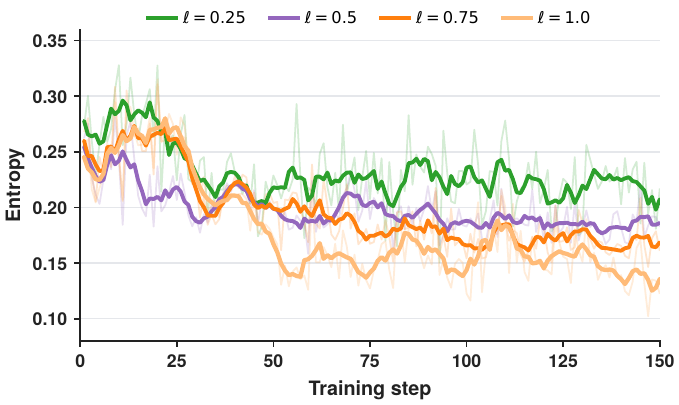}
    \caption{Actor entropy during training at the four support-switch percentiles on the Qwen3-1.7B student with the NTRL teacher.}
    \label{fig:entropy-percentile}
\end{wrapfigure}

\subsection{Sensitivity to the support-switch percentile}
\label{sec:ablate-hyper}
In \S\ref{sec:algo}, the support switches when overlap \(\rho\) crosses a percentile of the prefix overlaps, by default the median \(\rho_{0.5}\). This subsection measures how sensitive O$^{2}$PD is to that percentile. We keep \(\ell=0.2\), a reset atom, and overlap-gated \(\alpha\) fixed, and replace \(\rho_{0.5}\) by \(\rho_{0.25}\), \(\rho_{0.5}\), \(\rho_{0.75}\), or \(\rho_{1.0}\) (Tab.~\ref{tab:hyper-gate}). Tokens below the chosen percentile stay on \(T\cup\{r\}\); tokens at or above it use \(I\cup\{r\}\). We report Avg@8 and pass@8 on AIME24, AIME25, and HMMT25.
\begin{table}
\centering
\caption{Support-switch percentile on the Qwen3-1.7B (student) ← Qwen3-4B-NTRL (teacher).}
\label{tab:hyper-gate}
\small
\setlength{\tabcolsep}{2pt}
\renewcommand{\arraystretch}{0.9}
\setlength{\aboverulesep}{0.6ex}
\setlength{\belowrulesep}{0.4ex}
\begin{tabular}{@{}l cc cc cc c@{}}
\toprule
 & \multicolumn{2}{c}{AIME24} & \multicolumn{2}{c}{AIME25} & \multicolumn{2}{c}{HMMT25} & \\
\cmidrule(lr){2-3}\cmidrule(lr){4-5}\cmidrule(lr){6-7}
Percentile & Avg@8 & pass@8 & Avg@8 & pass@8 & Avg@8 & pass@8 & Avg \\
\midrule
\(0.25\) & 51.67 & 66.67 & 42.92 & 63.33 & 23.75 & 36.67 & 39.45 \\
\rowcolor{methodrow}
\(0.5\) & \textbf{54.17} & \textbf{73.33} & \textbf{45.83} & \textbf{70.00} & \textbf{25.83} & \textbf{43.33} & \textbf{41.94} \\
\(0.75\) & 52.50 & 70.00 & 45.42 & 66.67 & 25.42 & 40.00 & 41.11 \\
\(1.0\) & 53.75 & \textbf{73.33} & 43.33 & 66.67 & 24.58 & 40.00 & 40.55 \\
\bottomrule
\end{tabular}
\end{table}
The median is the best of the four percentiles. Percentile \(0.5\) has the highest Avg@8 on every benchmark. Switching to \(I\cup\{r\}\) at \(0.25\) hurts more than leaving tokens on \(T\cup\{r\}\) through \(0.75\) or \(1.0\).
Actor entropy declines over training at every percentile, and the decline is larger as the percentile approaches \(1.0\) (Fig.~\ref{fig:entropy-percentile}), where almost all prefix tokens remain on \(T\cup\{r\}\). That trend is consistent with leftover renormalization on \(T\cup\{r\}\) (Proposition~\ref{prop:face-renorm}).
We therefore use percentile \(0.5\): it yields the best performance of the four percentiles, and its entropy is not the lowest, so the student can keep exploration.
 
\section{Conclusion}
In this paper, we study two questions in OPD: \emph{which tokens to align}, and \emph{how to align those tokens}.
For \emph{which}, we find that teacher supervision is more reliable on the student prefix than on the suffix. For \emph{how}, we prove that minimizing $\RKL$ on the teacher top-$k$ neither covers teacher-only atoms at low overlap nor pins teacher mass on shared atoms at high overlap.
Therefore, we propose O$^{2}$PD, which concentrates on the alignment of the $20\%$ prefix of each rollout via a designed overlap-gated $\alpha$-divergence. 
The per-token overlap between teacher and student is used to adaptively modulate $\alpha$ and switch between Reset Support and Support Convergence.
On seven benchmarks O$^{2}$PD attains the best average performance among compared methods, improving over OPD by $+1.89$, $+4.81$, and $+5.29$ on the three different student--teacher pairs.

\subsection*{AI use statement}

In this work we used generative AI tools (LLM writing and coding assistants) for the following uses.

Writing. We used them to outline the paper, draft sections, edit for clarity, format \LaTeX{} tables and captions, and suggest titles and keywords. The authors reviewed and revised all submitted text.

Retrieval and discovery. We used them to search and summarize retrieved papers. Every citation was checked against a public source; unverified references were discarded.

Research ideation and execution. We used them to refine the research questions and conceptual framework, to formulate and check mathematical claims and proof sketches. They were not used to generate synthetic data.

\subsection*{Ethics statement}

The study distills language models on public math benchmarks and does not involve human subjects. Dataset licenses and intended-use constraints will be listed here before submission.

\bibliography{o2pd}
\bibliographystyle{o2pd}

\clearpage
\appendix
\setcounter{tocdepth}{2}
\startcontents[appendix]
\printcontents[appendix]{}{1}{\section*{Contents}\medskip}

\clearpage

\section{O$^{2}$PD training step}
\label{app:algo}

Algorithm~\ref{alg:o2pd} implements \eqref{eq:o2pd}. Only prefix tokens enter the loss. Overlap $\rho$ maps $\alpha$ from the prefix min and max, and switches Reset Support $T\cup\{r\}$ with Support Convergence $I\cup\{r\}$ at the median of prefix overlaps. Reported runs use $\ell=0.2$ and $k=16$ (\S\ref{sec:setup}).

\begin{algorithm}[h]
\caption{O$^{2}$PD update on one student rollout.}
\label{alg:o2pd}
\small
\begin{algorithmic}[1]
\Require student $\pi_s$, teacher $\pi_t$, prompt $x$, prefix fraction $\ell$, top-$k$ size $k$
\State sample $y\sim\pi_s(\cdot\mid x)$
\State $\mathcal{P}\leftarrow\{i:1\le i\le \ell\,|y|\}$
\For{$i\in\mathcal{P}$}
    \State $p^{(i)}\leftarrow\pi_t(\cdot\mid x,y_{<i})$,\ $q^{(i)}\leftarrow\pi_s(\cdot\mid x,y_{<i})$
    \State $T_i\leftarrow\TopK(p^{(i)})$,\ $S_i\leftarrow\TopK(q^{(i)})$,\ $I_i\leftarrow T_i\cap S_i$
    \State $\rho_i\leftarrow\lvert I_i\rvert/k$
\EndFor
\State $\mathrm{low}\leftarrow\min_{i\in\mathcal{P}}\rho_i$,\ $\mathrm{high}\leftarrow\max_{i\in\mathcal{P}}\rho_i$
\State $\rho_{0.5}\leftarrow$ median of $\{\rho_i:i\in\mathcal{P}\}$
\State $\mathcal{L}\leftarrow 0$
\For{$i\in\mathcal{P}$}
    \State $\mathrm{gate}\leftarrow\mathrm{clip}\bigl((\rho_i-\mathrm{low})/(\mathrm{high}-\mathrm{low}),0,1\bigr)$
    \State $\alpha_i\leftarrow 1-2\cdot\mathrm{gate}$
    \If{$\rho_i<\rho_{0.5}$}
        \State $A\leftarrow T_i\cup\{r\}$ \Comment{Reset Support}
        \State $\pi_p(a)\leftarrow p^{(i)}(a)$ for $a\in T_i$, $\pi_p(r)\leftarrow 1-p^{(i)}(T_i)$; likewise $\pi_q$ from $q^{(i)}$
    \Else
        \State $A\leftarrow I_i\cup\{r\}$ \Comment{Support Convergence}
        \State $\pi_p(a)\leftarrow p^{(i)}(a)$ for $a\in I_i$, $\pi_p(r)\leftarrow 1-p^{(i)}(I_i)$; likewise $\pi_q$ from $q^{(i)}$
    \EndIf
    \State $\mathcal{L}\leftarrow D_{\alpha_i}(\pi_q\Vert\pi_p)$
\EndFor
\State update $\pi_s$ on $\mathcal{L}$
\end{algorithmic}
\end{algorithm}

\section{Gradient proofs}
\label{app:proofs}

This section proves Lemma~\ref{lem:support-restriction} and Propositions~\ref{prop:covering}--\ref{prop:face-renorm}. Throughout, $q=\mathrm{softmax}(z)$ on $V$, $p$ does not depend on $z$, $T=\TopK(p)$, $S=\TopK(q)$, and $I=T\cap S$. We always assume the support in a display is nonempty and that the corresponding student mass is positive, so renormalized restrictions are well-defined. No further assumption is placed on the shapes of $p$ and $q$. The reset atom is introduced only in \S\ref{app:fkl}--\S\ref{app:face}; the failure statements in \S\ref{app:restriction}--\S\ref{app:sampled-min} concern $L_T$ without the reset atom.

\subsection{$\RKL$ simplex gradient}
\label{app:rkl-grad}

Let $\pi=\mathrm{softmax}(\zeta)$ on a finite simplex with strictly positive target $\pi_p$, and let $L=\KL(\pi\Vert\pi_p)=\sum_i\pi_i\log(\pi_i/\pi_{p,i})$. The softmax Jacobian is $\partial\pi_i/\partial\zeta_a=\pi_i(\delta_{ia}-\pi_a)$. Differentiating,
\begin{align}
\frac{\partial L}{\partial\zeta_a}
&=\sum_i\bigl(\log\tfrac{\pi_i}{\pi_{p,i}}+1\bigr)\,\pi_i(\delta_{ia}-\pi_a)
\nonumber\\
&=\pi_a\bigl(\log\tfrac{\pi_a}{\pi_{p,a}}+1\bigr)
-\pi_a\sum_i\pi_i\bigl(\log\tfrac{\pi_i}{\pi_{p,i}}+1\bigr)
\nonumber\\
&=\pi_a\bigl(\log\tfrac{\pi_a}{\pi_{p,a}}+1\bigr)-\pi_a(L+1)
=\pi_a\bigl(\log\tfrac{\pi_a}{\pi_{p,a}}-L\bigr).
\label{eq:app-rkl-grad}
\end{align}
The convention $0\log 0=0$ is used throughout. If $\pi_a\to 0$ while $L$ remains bounded, then $\pi_a\log\pi_a\to 0$, hence $\partial L/\partial\zeta_a\to 0$. Applying \eqref{eq:app-rkl-grad} on the simplex of $T$ with $\pi=q^{T}$ and $\pi_p=p^{T}$ yields
\begin{equation}
\label{eq:rkl-simplex-grad}
\frac{\partial L_T}{\partial \zeta_a}
=q^{T}_a\Bigl(\log\frac{q^{T}_a}{p^{T}_a}-L_T\Bigr),\qquad a\in T.
\end{equation}
The right-hand side vanishes whenever $q^{T}_a=0$. Membership $a\in T\setminus S$ does not force $q(a)=0$, but it does place $a$ outside $S$, so student mass on those atoms is typically far below the mass on $S$. Softmax never attains exact zeros; $q(T\setminus S)=0$ is a slice used in Proposition~\ref{prop:covering}, not a training constraint.

\subsection{Support restriction}
\label{app:restriction}

\begin{proof}[Proof of Lemma~\ref{lem:support-restriction}]
For $c\in T$,
\[
q^{T}_c=\frac{q(c)}{q(T)}=\frac{e^{z_c}}{\sum_{b\in T}e^{z_b}}.
\]
The partition function of $q$ over $V$ cancels, so $q^{T}$ depends on $z$ only through $\{z_b:b\in T\}$. The teacher restriction $p^{T}$ does not depend on $z$. Therefore $L_T=\sum_{a\in T}q^{T}_a\log(q^{T}_a/p^{T}_a)$ is a function of those logits alone, which gives $\partial L_T/\partial z_a=0$ for every $a\notin T$. Every $a\in S\setminus T$ and every $a\in V\setminus(T\cup S)$ falls in this set, so $\RKL$ on $T$ has no coordinate there.
\end{proof}

\subsection{Covering channel}
\label{app:covering}

\begin{proof}[Proof of Proposition~\ref{prop:covering}]
The assumption $q(T\setminus S)=0$ with $q(I)>0$ gives $q(T)=q(I)$ and $q^{T}_a=0$ for every $a\in T\setminus S$. Substituting into \eqref{eq:app-rkl-grad} yields $\partial L_T/\partial\zeta_a=0$ on $T\setminus S$. The remaining coordinates live on $I$, and
\[
L_T=\sum_{a\in I}q^{I}_a\log\frac{q^{I}_a}{p(a)/p(T)},
\]
which is $\RKL$ between $q^{I}$ and the restriction of $p^{T}$ to $I$. Lemma~\ref{lem:support-restriction} gives $\partial L_T/\partial z_a=0$ for $a\in S\setminus T$.

If instead $q(T\setminus S)$ is small but positive, \eqref{eq:app-rkl-grad} remains proportional to $q^{T}_a$ on $T\setminus S$. Emptiness of $T\setminus S$ as a set is not what shuts the covering channel. For $a\in T\setminus S$,
\[
\frac{\partial L_T}{\partial\zeta_a}=0
\quad\Longleftrightarrow\quad
q^{T}_a=0
\quad\text{or}\quad
\log\frac{q^{T}_a}{p^{T}_a}=L_T.
\]
Thus $a\notin S$ does not imply a vanishing gradient. The first-order covering share
\[
\rho(L_T)
:=\frac{\sum_{a\in T\setminus I}\bigl|\partial L_T/\partial\zeta_a\bigr|}
{\sum_{a\in T}\bigl|\partial L_T/\partial\zeta_a\bigr|}
\]
is zero whenever every $a\in T\setminus I$ satisfies the display above (and is undefined if the denominator is zero).
\end{proof}

\subsection{Renormalized match on $T$}
\label{app:sampled-min}

\begin{proof}[Proof of Proposition~\ref{prop:sampled-min}]
$L_T=\KL(q^{T}\Vert p^{T})$ is $\RKL$ on the finite simplex of $T$. Gibbs' inequality gives $\KL(\nu\Vert\mu)\ge 0$ with equality if and only if $\nu=\mu$, provided $\mu$ is strictly positive on that simplex. Softmax teachers assign $p(a)>0$ on $T$, so $p^{T}$ is strictly positive. The unique minimizer on the relative interior is therefore $q^{T}=p^{T}$. This identity is equivalent to $q(a)=c\,p(a)$ for $a\in T$ with the same $c=q(T)/p(T)$. It does not force $c=1$, and $L_T$ does not depend on logits outside $T$ (Lemma~\ref{lem:support-restriction}), so the minimizer is not $q=p$ on $V$.
\end{proof}

For completeness we record the face without a reset atom that is \emph{not} used as a training constraint. If $q(T\setminus S)=0$ and $q(I)=1$, then $q^{T}=q^{I}$ and
\[
L_T=\sum_{a\in I}q(a)\log\frac{q(a)}{p(a)/p(T)}
=-H(q)-\sum_{a\in I}q(a)\log p(a)+\log p(T).
\]
The last term does not depend on $q$. Minimizing over the simplex of $I$ gives $q(a)\propto p(a)$ on $I$, hence $q(a)=p(a)/p(I)$ for $a\in I$. Then $q(a)\ge p(a)$, with equality for a positive $p(a)$ if and only if $p(I)=1$. This characterizes the slice; it does not prove that SGD remains on the slice.

\subsection{FKL covering}
\label{app:fkl}

On a simplex with $\pi_q=\mathrm{softmax}(\zeta)$ and $L_F=\KL(\pi_p\Vert\pi_q)=\sum_i\pi_{p,i}\log(\pi_{p,i}/\pi_{q,i})$, the same Jacobian yields
\begin{align}
\frac{\partial L_F}{\partial\zeta_a}
&=\sum_i\pi_{p,i}\Bigl(-\frac{1}{\pi_{q,i}}\Bigr)\partial\pi_{q,i}/\partial\zeta_a
=-\sum_i\pi_{p,i}(\delta_{ia}-\pi_q(a))
=\pi_q(a)-\pi_p(a).
\label{eq:app-fkl-grad}
\end{align}

\begin{proof}[Proof of Proposition~\ref{prop:fkl-cover}]
The cancellation in Lemma~\ref{lem:support-restriction} shows that $\KL(p^{T}\Vert q^{T})$ depends on $z$ only through $\{z_a:a\in T\}$. Identity \eqref{eq:app-fkl-grad} on that simplex, or on $T\cup\{r\}$, is $\partial L_F/\partial\zeta_a=\pi_q(a)-\pi_p(a)$. If $a\in T\setminus S$, $\pi_q(a)=0$ and $\pi_p(a)>0$, then $\partial L_F/\partial\zeta_a=-\pi_p(a)<0$, so gradient descent raises $\zeta_a$.

If $p(T\setminus S)>0$ and $q^{T}|_{T\setminus S}\to 0$, the sum $\sum_{a\in T\setminus S}p^{T}_a\log(p^{T}_a/q^{T}_a)$ diverges to $+\infty$. For $\RKL$ the same limit is finite: $q^{T}$ remains a distribution on $I$ and $L_T$ reduces as in Proposition~\ref{prop:covering}.

On $T\cup\{r\}$, teacher reset mass is $p_r=1-p(T)$ and student reset mass is $q_r=1-q(T)$. Identity \eqref{eq:app-fkl-grad} on the reset coordinate is $\partial L_F/\partial\zeta_r=q_r-p_r$. Low overlap typically places substantial student mass outside $T$, so $q_r>p_r$ and descent lowers the reset coordinate, which is a pooled suppression of $S\setminus T$.
\end{proof}

\subsection{Leftover renormalization}
\label{app:face}

Proposition~\ref{prop:covering} keeps leftover $\RKL$ factors proportional to student mass, so $\varepsilon=q(T\setminus I)$ stays small rather than rising to $p(T\setminus I)$. A reset atom does not reopen that channel. The next statement compares $\RKL$ on $T\cup\{r\}$ and on $I\cup\{r\}$ with leftover coordinates frozen at a possibly positive $\varepsilon$; the uncovered face is $\varepsilon=0$.

\begin{proof}[Proof of Proposition~\ref{prop:face-renorm}]
The assumption $T\setminus I\neq\emptyset$ with $I=T\cap S$ gives $T\setminus I=T\setminus S$. Softmax teachers assign $p(a)>0$ on $T$, so $p(T\setminus I)>0$. Combined with $p(T)<1$,
\[
c:=p(I)+1-p(T)=1-p(T\setminus I)
\]
lies in $(0,1)$, and $p(I)=p(T)-p(T\setminus I)<1$.

On $T\cup\{r\}$, $\pi_p(a)=p(a)$ for $a\in T$ and $\pi_p(r)=1-p(T)$. Freeze $q$ on $T\setminus I$ with total mass $\varepsilon\in[0,1)$. The leftover $\RKL$ terms are then constant in the $I\cup\{r\}$ coordinates. Write $w$ for the remaining masses on $I\cup\{r\}$, so $w(a)=q(a)$ for $a\in I$, $w(r)=1-q(I)-\varepsilon$, and $\sum w=1-\varepsilon$. Let $m=(p|_I,\,1-p(T))$, hence $\sum m=c$, and
\[
L_{T,r}
=\sum_i w_i\log\frac{w_i}{m_i}
+C_{\varepsilon}
=(1-\varepsilon)\,\KL\Bigl(\frac{w}{1-\varepsilon}\Big\Vert\frac{m}{c}\Bigr)
-(1-\varepsilon)\log\frac{c}{1-\varepsilon}
+C_{\varepsilon},
\]
where $C_{\varepsilon}=\sum_{a\in T\setminus I}q(a)\log(q(a)/p(a))$ does not depend on $w$. Gibbs' inequality yields a unique minimizer $w=(1-\varepsilon)\,m/c$, i.e.
\[
q(a)=\frac{(1-\varepsilon)\,p(a)}{1-p(T\setminus I)}\ (a\in I),\qquad
1-q(I)-\varepsilon=\frac{(1-\varepsilon)\,(1-p(T))}{1-p(T\setminus I)}.
\]
Thus $q(a)/p(a)=(1-\varepsilon)/(1-p(T\setminus I))$ on $I$, so $q(a)>p(a)$ if and only if $\varepsilon<p(T\setminus I)$. Setting $\varepsilon=0$ recovers the uncovered-face formulae. As $\varepsilon\to 0$, the leftover contribution $C_{\varepsilon}\to 0$ because $x\log(x/p(a))\to 0$ for each $a\in T\setminus I$.

On $I\cup\{r\}$, leftover mass is part of the reset atom: $\pi_p(a)=p(a)$ on $I$ and $\pi_p(r)=1-p(I)$. This pair is a strictly positive probability. Gibbs' inequality gives the unique interior minimizer $q|_I=p|_I$ and $1-q(I)=1-p(I)$. The coordinate $T\setminus I$ is not in the simplex over $I\cup\{r\}$, so the $I$-target does not depend on $\varepsilon$. If leftover mass is additionally frozen at $\varepsilon\le 1-p(I)$, that interior point remains feasible by placing the remaining reset mass on $V\setminus T$. When $T=I$ the two simplices coincide, and the unconstrained interior of $T\cup\{r\}$ is $q|_T=p|_T$, which requires $\varepsilon=p(T\setminus I)$.

The simplex gradient on $I\cup\{r\}$ has only two kinds of coordinates:
\begin{equation}
\label{eq:Ir-grad}
\frac{\partial L_{I,r}}{\partial\zeta_a}
=q(a)\Bigl(\log\frac{q(a)}{p(a)}-L_{I,r}\Bigr)\ (a\in I),
\qquad
\frac{\partial L_{I,r}}{\partial\zeta_r}
=(1-q(I))\Bigl(\log\frac{1-q(I)}{1-p(I)}-L_{I,r}\Bigr).
\end{equation}\nonumber
Identity \eqref{eq:Ir-grad} is \eqref{eq:app-rkl-grad} on $I\cup\{r\}$. For vocabulary logits $z$, the simplex logits may be realized as $\zeta_a=z_a$ for $a\in I$ and $\zeta_r=\mathrm{logsumexp}(z|_{V\setminus I})$. Then $\partial L_{I,r}/\partial z_c=q(c)(\log(q(c)/p(c))-L_{I,r})$ for $c\in I$, and $\partial L_{I,r}/\partial z_c=q(c)(\log((1-q(I))/(1-p(I)))-L_{I,r})$ for $c\notin I$. Tokenwise $\log(q/p)$ therefore appears only on $I$. The analogous identities on $T\cup\{r\}$ place tokenwise $\log(q/p)$ on every $a\in T$, including $T\setminus S$.
\end{proof}

\begin{lemma}[Gradient coincidence]
\label{lem:coincide}
If $q(T\setminus I)=0$ and $q(I)>0$, then $L_T=L_I+\log\bigl(p(T)/p(I)\bigr)$. The second term does not depend on $z$, so $L_T$ and $L_I$ induce the same student-logit gradient. In particular both first-order covering shares are zero whenever they are defined.
\end{lemma}

\begin{proof}
The assumption gives $q(T)=q(I)$ and $q^{T}_j=q^{I}_j$ for $j\in I$, together with $q^{T}|_{T\setminus I}=0$. Then
\begin{align*}
L_T
&=\sum_{j\in I}q^{I}_j\log\frac{q^{I}_j}{p_j/p(T)}
=\sum_{j\in I}q^{I}_j\log\frac{q^{I}_j}{p_j}+\log p(T),\\
L_I
&=\sum_{j\in I}q^{I}_j\log\frac{q^{I}_j}{p_j/p(I)}
=\sum_{j\in I}q^{I}_j\log\frac{q^{I}_j}{p_j}+\log p(I).
\end{align*}
Subtracting yields $L_T=L_I+\log(p(T)/p(I))$. For $a\in T\setminus I$ the $\RKL$ identity \eqref{eq:app-rkl-grad} is zero because $q^{T}_a=0$. Gradients on $I$ therefore coincide, and the covering share of $L_T$ is zero.
\end{proof}

Lemma~\ref{lem:coincide} is the boundary case of Proposition~\ref{prop:covering} for $\RKL$ on $T$. Leftover renormalization (Proposition~\ref{prop:face-renorm}) does not degenerate: the $T\cup\{r\}$ $I$-target equals $p|_I$ only when $\varepsilon=p(T\setminus I)$, which is the unconstrained interior rather than the small-$\varepsilon$ regime.

\section{Endpoints of $\alpha$-divergence}
\label{app:alpha-lim}

This appendix verifies the two KL limits of \eqref{eq:alpha-div} used in \S\ref{sec:algo}. The family is due to Amari~\cite{amari2009alpha}; the calculation is for the displayed coefficients and argument order. Throughout, $A$ is finite and $\pi_q(i)>0$, $\pi_p(i)>0$ for every $i\in A$.

Write
\[
D_{\alpha}(\pi_q\Vert\pi_p)
=\frac{4}{1-\alpha^{2}}\Biggl(1-\sum_{i\in A}\pi_q(i)^{\frac{1-\alpha}{2}}\pi_p(i)^{\frac{1+\alpha}{2}}\Biggr),
\qquad \alpha\in(-1,1).
\]

\textbf{$\alpha\to 1$.}
Set $\beta=(1-\alpha)/2$, so $\beta\to 0^+$ and $\alpha=1-2\beta$. Then $1-\alpha^{2}=4\beta(1-\beta)$ and $4/(1-\alpha^{2})=1/(\beta(1-\beta))$. The inner sum is
\[
\sum_{i\in A}\pi_p(i)\Bigl(\frac{\pi_q(i)}{\pi_p(i)}\Bigr)^{\beta}
=\sum_{i\in A}\pi_p(i)\,e^{\beta\log(\pi_q(i)/\pi_p(i))}
=1+\beta\sum_{i\in A}\pi_p(i)\log\frac{\pi_q(i)}{\pi_p(i)}+O(\beta^{2}).
\]
Hence
\[
1-\sum_{i\in A}\pi_q(i)^{\beta}\pi_p(i)^{1-\beta}
=\beta\,\KL(\pi_p\Vert\pi_q)+O(\beta^{2}),
\]
and $D_{\alpha}=(\beta\,\KL(\pi_p\Vert\pi_q)+O(\beta^{2}))/(\beta(1-\beta))\to\KL(\pi_p\Vert\pi_q)$.

\textbf{$\alpha\to -1$.}
Set $\gamma=(1+\alpha)/2$, so $\gamma\to 0^+$ and $\alpha=2\gamma-1$. Then $1-\alpha^{2}=4\gamma(1-\gamma)$ and $4/(1-\alpha^{2})=1/(\gamma(1-\gamma))$. The inner sum is
\[
\sum_{i\in A}\pi_q(i)\Bigl(\frac{\pi_p(i)}{\pi_q(i)}\Bigr)^{\gamma}
=1+\gamma\sum_{i\in A}\pi_q(i)\log\frac{\pi_p(i)}{\pi_q(i)}+O(\gamma^{2}).
\]
Hence
\[
1-\sum_{i\in A}\pi_q(i)^{1-\gamma}\pi_p(i)^{\gamma}
=\gamma\,\KL(\pi_q\Vert\pi_p)+O(\gamma^{2}),
\]
and $D_{\alpha}\to\KL(\pi_q\Vert\pi_p)$.

\section{The two failures after O$^{2}$PD}
\label{sec:ablate-repair}

\begin{figure}[t]
    \centering
    \includegraphics[width=\linewidth]{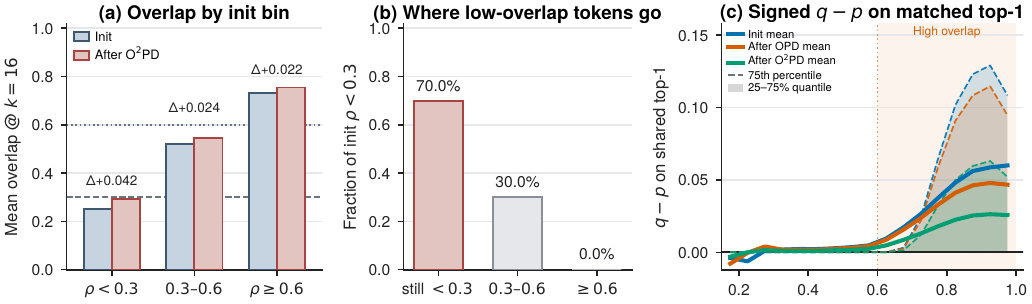}
    \caption{Prefix tokens at $k=16$ before and after O$^{2}$PD. (a)~Mean overlap by initial-overlap bin. (b)~Destination of tokens with initial $\rho<0.3$. (c)~Signed $q-p$ on the shared top-$1$.}
    \label{fig:repair}
\end{figure}

To test whether O$^{2}$PD relieves the two failures of top-$k$ OPD in \S\ref{sec:failures}, we follow prefix tokens at $k=16$ (Fig.~\ref{fig:repair}). Teacher-support Uncovering is read from mean overlap by initial-overlap bin and from the destination of tokens with initial $\rho<0.3$. Dominant-Mode Mismatch is read from the signed gap $q-p$ on the shared top-$1$.

Mean overlap rises in every bin, by $0.042$ for $\rho<0.3$, $0.024$ for $0.3$--$0.6$, and $0.022$ for $\rho\ge 0.6$. Of the tokens that start below $0.3$, $30\%$ move into $0.3$--$0.6$ and none reach $\ge 0.6$; $70\%$ remain below $0.3$. Under top-$k$ OPD, $87.3\%$ of that cohort remain below $0.3$ and none reach $\ge 0.6$. O$^{2}$PD moves more of the low-overlap cohort out of the lowest bin, and most of it is still there.
On tokens that share the teacher's top-$1$, the signed gap $q-p$ after O$^{2}$PD lies below the gap after top-$k$ OPD and remains positive at high overlap.
O$^{2}$PD therefore relieves Teacher-support Uncovering and Dominant-Mode Mismatch problems relative to top-$k$ OPD.

\section{Training configuration}
\label{app:hyper}

We train O$^{2}$PD in VERL~\cite{sheng2025hybridflow} on 16-NPU Ascend 910C machines. Chat templates disable thinking. The update is the distillation loss alone. Teacher log-probabilities are computed on the student prompt. Table~\ref{tab:train-pairs} lists the three student--teacher pairs. DSR1-1.5B is DeepSeek-R1-Distill-Qwen-1.5B, and the NTRL teacher is Qwen3-4B-Non-Thinking-RL. All runs use DAPO-MATH-17K with no additional filtering, right truncation, and data-loader seed $42$. Table~\ref{tab:train-shared} records the optimizer and rollout recipe. Table~\ref{tab:train-o2pd} records the O$^{2}$PD loss settings used in the reported runs.

\begin{table}[t]
\centering
\caption{Student--teacher pairs used for O$^{2}$PD.}
\label{tab:train-pairs}
\small
\begin{tabular}{ll}
\toprule
Student & Teacher \\
\midrule
Qwen3-1.7B & Qwen3-4B \\
DSR1-1.5B & JustRL-1.5B \\
Qwen3-1.7B & Qwen3-4B-NTRL \\
\bottomrule
\end{tabular}
\end{table}

\begin{table}[!t]
\centering
\caption{Shared training recipe for O$^{2}$PD.}
\label{tab:train-shared}
\small
\begin{tabular}{ll}
\toprule
Setting & Value \\
\midrule
Hardware & 16$\times$ Ascend 910C \\
Data & DAPO-MATH-17K, unfiltered \\
Optimizer & AdamW, $\beta=(0.9,0.999)$, weight decay $0.01$ \\
Learning rate & $2\times 10^{-6}$, cosine decay, min ratio $0.1$ \\
Warmup / grad clip & $10$ steps / $1.0$ \\
Train / mini-batch & $64$ / $16$ \\
Micro-batch per NPU & $4$ \\
Max prompt / response & $1536$ / $4096$ \\
Rollout $n$ / temperature & $1$ / $1.0$ \\
Precision & bfloat16 rollout, FSDP param and optimizer offload \\
KL loss & off \\
Loss clamps & max $10.0$, $\log q$ min $-10.0$ \\
Thinking & off \\
\bottomrule
\end{tabular}
\end{table}

\begin{table}[t]
\centering
\caption{O$^{2}$PD loss settings.}
\label{tab:train-o2pd}
\small
\begin{tabular}{ll}
\toprule
Setting & Value \\
\midrule
Top-$k$ size $k$ & $16$ \\
Prefix fraction $\ell$ & $0.2$ \\
Reset atom & on \\
$\alpha$ & mapped from prefix overlap min and max \\
Support switch & median of prefix overlap: $T\cup\{r\}$ below, $I\cup\{r\}$ at or above \\
Scheduled steps & $200$ (Qwen3-4B teacher); $300$ (NTRL and DSR1 pairs) \\
\bottomrule
\end{tabular}
\end{table}

O$^{2}$PD on the NTRL and DSR1 pairs was scheduled for $300$ steps, and O$^{2}$PD with the Qwen3-4B teacher for $200$ steps. Downstream scores use intermediate checkpoints at step $60$ or $100$, not necessarily the last scheduled step.

\section{Evaluation}
\label{app:eval}

Downstream evaluation uses the same decoding and grading pipeline for distilled checkpoints and reference models. The cross-prefix protocol in \S\ref{sec:where} reports pass@1 on MATH500 and is separate from the settings below.

Every model is decoded with vLLM. Thinking mode is disabled. All problems use the model's chat template with the user prompt:
\begin{quote}
\small
\texttt{\{problem\} Please reason step by step, and put your final answer within \textbackslash boxed\{\}.}
\end{quote}
Decoding uses temperature $0.6$, top-$p=0.95$, sampling top-$k=20$, and a maximum of $31,744$ tokens.

A response is graded by extracting the final \texttt{\textbackslash boxed\{\}} expression and checking mathematical equivalence against the reference answer with a rule-based verifier. The verifier applies symbolic-equivalence normalization for fractions, radicals, and other equivalent forms. A response without a parseable boxed answer is counted as incorrect.

\end{document}